\documentclass{article}
\usepackage{titlesec}
\usepackage{booktabs}
\titleclass{\subsubsubsection}{straight}[\subsection]
\newcounter{subsubsubsection}[subsubsection]
\renewcommand\thesubsubsubsection{\thesubsubsection.\arabic{subsubsubsection}}
\titleformat{\subsubsubsection}{\normalfont\normalsize\bfseries}{\thesubsubsubsection}{1em}{}
\titlespacing*{\subsubsubsection}
{0pt}{3.25ex plus 1ex minus .2ex}{1.5ex plus .2ex}

\usepackage[english]{babel}
\usepackage{times}

\usepackage[letterpaper,top=2cm,bottom=2cm,left=3cm,right=3cm,marginparwidth=1.75cm]{geometry}
\usepackage{comments}
\usepackage{ulem}

\usepackage{natbib}
\usepackage{soul}

\title{Computational modeling of semantic change}
\author{Nina Tahmasebi$^1$, Haim Dubossarsky$^2$\\$^1$ {\small University of Gothenburg, Sweden, nina.tahmasebi@gu.se} \\ $^2$ {\small Queen Mary University of London \& University of Cambridge, UK, h.dubossarsky@qmul.ac.uk}}
\date{}
\begin{document}
\maketitle

\begin{abstract}

In this chapter\footnote{This chapter is submitted to Routledge Handbook of Historical Linguistics, 2nd Edition} we  provide an overview of computational modeling for semantic change using large and semi-large textual corpora. We aim to provide a key for the interpretation of relevant methods and evaluation
techniques, and also provide insights into important aspects of the computational study of semantic change. 
We discuss the pros and cons of different classes of models with respect to the properties of the data from which one wishes to model semantic change, and which avenues are available to evaluate the results. 
\end{abstract}

\section{Introduction}
Semantic change concerns change at the meaning level of a word. Examples are plentiful, and concern rapid change as well as slow, gradual change. 
Some changes are permanent, while others have a limited lifespan and disappear after having trended for a time. Regardless, changes in word meaning affect our language and the way we use our words to express ourselves. What is more, semantic change has ripple effects into lexical change; when one word gains a new sense, another word may become obsolete, or they may split the semantic space between them \citep{LEHRER+1985+283+296, anttila1989historical, xu2015computational}. For example, the words \textit{skarp} and \textit{vass} in Swedish are two synonyms for the English word ``sharp'' that take different objects, like \textit{skarp syn} ``sharp eye sight'' and \textit{vass kniv} ``sharp knife''. 

Changes in our language often reflect changes in our everyday lives, societies, and culture. There are thus many reasons why the study of semantic change is important. 
With the introduction of powerful computers and the increasing availability of digital, diachronic texts, we have seen the rise of the computational study of semantic change in the past two decades. 

Traditional research into semantic change lists many categories of  change, as well as reasons for their occurrence, but in this chapter, we will not (a) explain what semantic change is,  (b) explain how it can be divided into categories, or (c) dwell on the reasons why it occurs. Those topics are addressed in the excellent chapter on semantic change in this volume. The focus on this chapter is on computational modeling of semantic change. While semantic change is often studied over long time spans, even ephemeral semantic changes are important objects of study. 
We are often unaware of all the ongoing changes in domains where we typically do not operate. For example, youth slang is unknown territory for most people (who are either out of the age range or without direct contact with someone in the age range). For this reason, we will consider both long- and short-term semantic changes in this chapter. 

Work in the field of computational linguistics is motivated both by the desire to provide data-driven linguistic knowledge and by the desire to apply an interesting modeling tool to a complex, temporally evolving data type. The work may also be motivated by an interest in temporal  information retrieval, and often by its potential as a computational tool for lexicography. 
Recently, we have   also been seeing work that applies computational modeling of semantic change to problems in other non-linguistic text-based research fields, such as history \citep{BizzoniDegaetanoOrtliebFankhauseretal-2020}, medicine \citep{vylomova-etal-2019-evaluation}, and political science \citep{marjanen2019clustering, tripodi-etal-2019-tracing}.

Detecting semantic change often relies on methods that automatically encode meaning from raw text, that is, they are based on representations of meaning. The models currently used to encode meaning for studying semantic change are \textbf{unsupervised} and do not make use of preexisting knowledge bases (such as dictionaries or lexicons) nor do they involve any human input. The basis for our work is thus textual data and a computational method to extract information about words and their meaning from the text and so create meaning representations, 
on the basis of which we can detect semantic change. It is thus extremely important how we extract meaning from text, and how we choose to represent it. 
Current models of extracting meaning from text work on three different levels: the token, sense, and type levels. The first level (token) is that of individual word occurrences. These occurrences are often called contexts and are defined either as a fixed number of words around a target word, or a sentence, or other larger text unit, in which the target word occurs. Most active research is currently being carried out at this level. 
The second level is that of sense, where we either try to directly model a word's senses, for example with the help of word sense induction or topic modeling, or group token-level representations into sense-level representations. There is ongoing work on the latter, but there is currently little active research on modeling senses directly. We will therefore not discuss this level in much detail in this chapter. 
Finally, the third level is the word level (type) representations. This level collapses all information about a word into one, and results in an average representation of all the meanings of a word that appear in different contexts. 
The first major push of computational semantic change research was done using type-based models, and only later were token and sense level models of meaning revisited. 

\bigskip\noindent\textbf{What does it mean to model semantic change computationally? } \label{subsec:researchformulations}
The answer depends on which question one sets out to answer. When it comes to semantic change research, there are three main research formulations:  

\begin{enumerate}
    \item The first research formulation relates to \textbf{modeling  semantic change as it is  exhibited in a specific corpus}.  We are interested in what happens to a word as used in a specific corpus, as opposed to its use in general; thus, we limit the textual material that we use. We allow semantic change to be found only on the words present in this material, and accept only those changes that are apparent in this specific material. The fact that a word may have changed differently outside of the corpus does not matter. 
    \item
The second research formulation is significantly broader and aims to model language in general: \textbf{Have our words changed their meaning?} We study this by using different textual sources as a proxy for the world around us and language as a \textit{whole}. This approach is what is generally understood when we speak of studying lexical semantic change. We want to contribute to the general knowledge of our language, to dictionaries, historical linguistic studies, or to specify laws of semantic change (whether language specific or language agnostic).
\item A  third research formulation (so far hypothetical) can be seen as a middle path between the two above. This formulation concerns contrasting the language in one corpus with contemporary general language: Does a specific author, or community, exhibit a different language use from the general population? 
\end{enumerate}

In this chapter, we aim to provide a key for the interpretation of relevant methods, studies, and evaluation techniques, as well as provide insights into important aspects of the computational study of semantic change. 
We will focus on computational modeling of semantic change from large and semi-large textual corpora. 
Most of the work done in this field so far can be considered as a proof of concept, showing that computational methods can be indeed applied to corpora to automatically detect semantic change.  Until recently, there have been no common standards for how to choose and preprocess data, perform experiments, or evaluate and present results. 
However, researchers are slowly reaching consensus on some 
best practices for modeling, data collection, experimentation, and evaluation. The time has come to take larger steps towards data-driven linguistic inquiry, and to use computational modeling of semantic change to model societal and cultural change.

\section{Computational models of meaning}

To date, several classes of methods with different ways of modeling word meaning have been used to tackle semantic change detection. 
These classes follow the general trend in computational linguistics and natural language processing (NLP) and range from word sense induction-based (clustering) methods \citep{tahmasebi2013models, mitra2015automatic}, to topic models \citep{lau2014learning, frermann2016bayesian} and, most prominently now, 
distributional semantic methods.  These methods operate at word level (all information about a word is compressed into one representation), sense level (where the information about a word is divided between different senses), or token level (all information about a word is divided into representations of each time it is used). 

The focus of this section will be on distributional semantic models that can be divided into three separate groups: (1) explicit count-based models (Section \ref{sec:countbased}), (2) static embedding models (Section \ref{sec:staticemb}) and, (3) contextualized embedding models (Section \ref{sec:contemb}).
The aim of distributional semantic models is to represent each word with one or several vectors, such that the vectors are distributed in the semantic vector space. We call these \textit{vector representations}, or just \textit{representations} for each word. Because words are represented as vectors, we can apply vector arithmetic, that is, we can add and subtract vectors from each other. This property allows us to investigate relations between words such as the relation between countries and their capitals or between singular and plural forms of words. However, the most important property of these models is that they enable us to capture the meaning of a word by operationalizing the distributional hypothesis. 

\subsection{Theoretical background}
Over the last 15 years there has been an immense and growing interest and research focus on developing models of computational meaning. These models differ significantly in terms of architecture (e.g., type of neural network design used, if any), computational complexity (i.e., what kind of \textit{information} the model is trying to learn) and scope (e.g., how much textual data is used for training). These models represent significant technological advances. However, despite huge differences between different models of computational meaning, they all rely on the same theoretical foundation, namely the distributional hypothesis (henceforth DH).

The DH suggests that we can infer the meaning of words by systematically analyzing the context in which they are used. For example, if we do not know the word \emph{cat}, or what object it signifies in the real world, we can form  a pretty good guess about its meaning when we read all the sentences (i.e., context of usage) in which the word \textit{cat} appears: ``a cat is meowing'', ``the cat drinks milk'', ``I took the cat to the vet''. Despite its relative simplicity, the DH provides the basis of the Artificial Intelligence  revolution we have seen in the last decade, by enabling computational models to carry out more and more complicated tasks like machine translation, dialog systems, and machine summarization that previously required human-level text understanding.

The DH was advanced almost simultaneously during the 1950s by Zellig Harris, an American semanticist and mathematician who argued that ``similarity of meaning equals similarity of usage'' (1954), and by the British linguist John F. Firth who coined the more famous phrase ``You shall know a word by the company it keeps.'' Importantly, DH has its roots in Saussurean Linguistic Structuralism, and is  inline with psycholinguistic accounts of how humans learn and represent meaning \citep{Tahmasebi2024, Brunila2022}.

For many years it was nearly impossible to model meaning based on the DH because correctly establishing  a word's usage pattern required huge amounts of text to be processed (i.e.,  a large and representative sample of the word's usage contexts, multiplied by the number of words in the vocabulary). In recent years however, the size of digitized textual resources has increased exponentially (and now also include many more languages). At the same time there have been unprecedented leaps in computational power and algorithms that process text in huge quantities. This combination has driven remarkable advancements in DH-inspired computational models of meaning described in this chapter.

\subsection{Count-based models of meaning}\label{sec:countbased}
The first DH inspired models were straightforward and based on the idea of counting frequencies of occurrence. For each \textit{target word}, the models count how many times other words appeared (co-occurred) in the \textit{context} of the target word in a large textual corpus. A context contains other words in close proximity to the target word. It can be a fixed number of words before and after the target word (a window), the whole sentence, or the whole paragraph in which the target word occurs. After having processed all target words in a corpus (typically, target words are chosen as the \textit{K} most frequent nouns or verbs), the result is a list of \textit{n} numbers per target word \textit{w}, mathematically referred to as a \textit{vector of n dimensions}. Each element in the list/vector (i.e., element $ i \in 1\ldots, n$) represents the number of times a  word \textit{i} appears in the context of the target word \textit{w}. Each list/vector of numbers represents a word's meaning.

It is easy to show how these vector representations, ultimately one vector for each word, capture meaning in accordance with the DH. Let us consider, for example, the similarity between the vectors of ``cat'' and ``dog''. The co-occurrence vectors of these two words will be very similar with respect to certain elements in the vectors, because the same words tend to co-occur with both cat and dog (e.g., ``play'', ``pet'', or ``veterinarian''). They are clearly quite different from other words like ``car'' (``speed'', ``gear'', ``garage''), or ``pray'' (``god'', ``worship'',  ``service'', ``rabbi'', ``mosque''). 
These count-based models can be used to create representations of a word as a whole (as described above), that is, type representations (see Table \ref{tab:co-occur} for an illustrative example).\footnote{Count-based models can be used to create token representations as well, see \cite{Schutze1998} for an example, but these have not been explored in semantic change research thus far. 
} In this case, the vector is a first order co-occurrence vector which means that each element in the vector corresponds to how often that word co-occurs with the target word in the corpus. This representation of the target word conflates all the different meanings, and can be seen as some sort of an average meaning representation where the sense that occurs more frequently has a higher effect on the representation of the word.

\begin{table}

\centering
\caption{Illustration for a co-occurance matrix with rows representing target words, and columns representing context words. Numerical value represent raw counts.}
\label{tab:co-occur}
\scalebox{0.8}{
     \begin{tabular}{|c|c|c|c|c|c|c|c|c|c|c|c|c|c|c|c|c|}
    \hline
            ~ & and & for & garage & gear & god & mosque & of & pet & play & rabbi & service & speed & that & the & vet & worship \\ \hline
        Cat & 114 & 58 & ~ & ~ & ~ & ~ & 757 & 451 & 224  & ~ & ~ & ~ & 424 & ~ & 120 & ~ \\ \hline
        Dog & 98 & 44 & 5 & ~ & ~ & ~ & 495 & 890 & 450 & ~ & ~ & 21 & 859 & 1434 & 90 & ~ \\ \hline
        Car & 265 & 78 & 179 & 67 & ~ & ~ & 46 & 1 & 18 &  ~ & 90 & 157 & 650 & 990 & ~ & 1 \\ \hline
        Pray & 89 & 22 & ~ & ~ & 170 & 45 & 700 & 2 & 10 & 117  & 240 & ~ & 56 & 1930 & ~ & 148 \\ \hline
    \end{tabular}
  }

    \label{task1_combo}
\end{table}

\subsubsection{Statistical processing of co-occurring patterns}
Critical examination of the data (see Table \ref{tab:co-occur}) shows that non-content words (e.g., stop words, function words, and prepositions) have high co-occurrence counts with almost all the words in the vocabulary. For example, words like ``the'', ``of'', ``for'', ``and'', which are among the most frequent words in English appear often with most of the words in the lexicon. However, these high frequency words contribute very little to the meaning of any word they co-occur with.

Statistical independence scores are used to counter the effect of the most frequent words in the language which, as said, are rather meaningless in the lexical sense. These scores enable capturing meaningful information from the co-occurrence patterns, rather than from the raw counts.\footnote{Stop words can often be filtered out using a simple stop word list that is readily available. However, PMI methods are used to handle very common words that are not typically found in a stop words lists.} One of the first methods developed was PMI (PointWise Mutual Information). PMI answers the following question: For a given pair of co-occurring words, how often do they appear together (in the context of one another) relative to their overall frequency? If, for example, ``cat'' has a general probability of .01\% in the entire text, and a joint probability of only .011\% to co-occur with the word ``the'', then the two words are not strongly associated (i.e., the appearance of ``the'' in the text is not a good indication that ``cat'' will follow it). On the other hand, if ``meow'' has a general probability of 0.005\% in the text, but a 0.1\% joint probability to appear together with ``cat'', then it is safe to (statistically) assume that the two are strongly associated and that when we find ``meow'', we will often also find ``cat''. 

Several variants of PMI exist, as well as other statistical methods to evaluate the ``importance'' of words (e.g., PPMI, tf-idf). However, as they are very similar in spirit to PMI, and we will not discuss them here. Although more sophisticated methods for vector meaning representation have been developed (see next subsections), PMI and PPMI still remain popular choices for research because of their simplicity and their interpretable representations -- the strength of the association between each word and its context words is explicitly represented in the vector representation.

\subsection{Predictive models}

Predictive models, to which static and contextual embeddings belong, are a group of machine learning algorithms designed to carry out a particular task without explicitly being directed how to do so (i.e., they are not instructed to count words that co-occur with other words). Such models are typically trained on a very large set of examples which in our case stem from very large textual corpora. These models learn the meaning of words directly from text, by training to predict a MASKED word given its context. 

\bigskip
[1] \textit{The cat} \textless MASK\textgreater{} \textit{after I accidentally stepped on his tail.}
\bigskip

In the above example, a predictive model will likely predict \textit{mewed} in place of the masked word. This training task is called a masked language modeling task and is common to both static and contextualized predictive models.\footnote{Despite its simplicity, it is a ubiquitous task, and most NLP models are trained on it (or on its variants), including Large Language Models.}

\subsubsection{Geometrical interpretation}\label{sec:geom_interp}

Predictive models, like count-based ones, aim to create a vector space in which words are represented as vectors and grouped according to meaning. These spaces, however, unlike the count-based ones, do not have interpretable dimensions.  We call these \textit{semantic spaces} because they are occupied by vectors that represent word meanings.

We can think of the vector space as a very large room.\footnote{A physical room has only three dimensions whereas our spaces are typically 100 to 200 dimensions large. Also our imaginary room has walls, that is boundaries, which our space does not. Nonetheless, the room analogy can be used for illustrative purposes} We aim to group words that are semantically similar close together in this room. 
This grouping is a simple reflection of  DH principles in a geometric space: words with similar usage context have  similar vectors that end up close together in the vector space. 
For example, the model aims to get all words relating to food in one region,  technical devices in another, and animals in a  third region. The grouping of words in such a vector space tells us that \textit{cat} is more similar to \textit{dog} than to \textit{pasta}, because \textit{cat} and \textit{dog} are located in the same region, while \textit{cat} and \textit{pasta} are, for example, in opposite corners. Often, the models also learn certain relations between categories of words, for example that \textit{cat} and \textit{cats} have the same relative position to each other, as other singular-plural pairs like \textit{chair} and \textit{chairs}. We can measure word similarity by the distance between vectors. We can use well-known distance metrics (like cosine distance or Euclidean distance) to determine their similarities. The smaller the distance (i.e., the closer the vectors are in the room), the more similar the vectors, and hence the meanings they represent.

When we start creating a vector space, we do not assume any advance knowledge of what words mean, and which words are similar to one another. We start the process by throwing all the words into the room at random, and then we incrementally train our model. Each time a target word \textit{w} is encountered in a sentence, the vector corresponding to \textit{w} (i.e., the vector that represents \textit{w}'s meaning) takes a small step in the direction suggested by the context words. How large 
a step the vector takes in the semantic space, or in which direction it moves, is determined by how good (or bad) the model was at predicting the correct word in the current masked-language iteration. If the model errs by a large amount, the size of the step, that is, the correction to its position in the semantic space, is larger than if the model is more accurate. 

Because of this random initialization of the model, some words have to travel only a short distance (as they happen to have started close to their final region), while other words need to travel a longer distance to reach their correct region. For a word to travel through the semantic space, it needs to take sufficiently many steps to the region where its semantically similar friends belong. A step is only taken if a word occurs in the corpus. Therefore,  if a word does not appear frequently enough, it will stop short on its journey and will not reach its correct region. If too many words remain in the wrong region when training stops, then the space becomes useless. That is, we cannot trust that word representations are accurate and hence that \textit{cat} is more similar to \textit{dog} than to \textit{cutlery} or that the relation between singular and plural is preserved for most nouns. On the other hand, if the process ends successfully, and a sufficient number of words end up in the correct region, we say that the model has \textit{converged}.  

It is important to note that there is no fixed place where  semantically similar words, for example \textit{foods} or \textit{animals}, should always end up. Due to the probabilistic nature of the methods, the words will end up in different regions each time the algorithm is run. It is therefore the relative position and distance between words in the semantic space that matters.                   This property is important  when one wants to compare words using vectors trained over different periods of time. This point will be addressed in Section \ref{ChoiceOfComputataionModel}.

\subsubsection{Static embedding models}\label{sec:staticemb}
Static embedding models are a suite of models whose starting point is a huge set of vectors, one per word. However, unlike the count-based vectors, each element in the vector corresponds to a learned dimension rather than another word, and we decide the length of the vectors (typically 50--300 dimensions) without regard to the size of the vocabulary. These vectors are responsible for two things: 1) they represent the ``internal states'' of the learning algorithm that determine how the model reacts to a given input; and 2) they comprise the numerical basis of the geometrical interpretation we gave above. 

Throughout the training process, a model tunes the numbers within each vector, which are called parameters (these are different from the hyper-parameters described in Section \ref{sec:hyperparam}) so that they will produce the right output for a certain input. For example, for an input of [1] above, the model would output ``mewed''. The training process itself is very similar to the way we tune a musical instrument -- we press a piano key or pluck a string on a violin, and then adjust the string's tension (up or down) to produce a more accurate sound in our next attempt. Predictive models learn the parameters in a very similar way. For every iteration in which they predict a masked word (in the above example ``mewed'', ``cried'', ``yelled'' would be reasonable predictions) they test whether their prediction was correct, and if not, update the parameters accordingly. 

The unique design of static models, with a strict mapping of \textit{one vector per word}, guarantees that the parameters related to a specific word are only updated if that particular word appears in a sentence, and remain unchanged when updating the parameters of other words. Ultimately, these learned parameters constitute the vector representations of the predictive models that we use to describe the meaning of words. 

\subsubsection{Contextual embedding models}\label{sec:contemb}
Contextualized embedding models, unlike static models, have no strict mapping between a word and its parameters (i.e., no one vector per word). During training, all the  parameters of the model are updated (in a very similar manner to the process described above). After training, the parameters of the model are dynamically 
 used to describe (and predict) a word in each context it appears in. 
 Consequently, contextualized models produce  \textit{a token vector, for every instance of a word} in the text, on the basis of the context words (typically in the sentence where the word occurs). As a result, a word is not represented by one vector in the semantic space but by many (as many vectors as there are instances of the word in the text).  Therefore, token-based representation improves our chances of deriving sense information because words that have multiple senses will have token vectors in different regions.  It means that, at least in theory, the vectors representing all instances of the word can be separated into classes (e.g. by means of clustering) where each class represents one word sense. 

Contextualized models are considered more expressive than static models for two main reasons. The first is their larger number of parameters (from 10 to a 1,000-fold more parameters) compared to static embeddings. The second is the design of the underlying neural network. Contextualized models have a funnel-like structure that looks like layers (cf. to static models that consist of a single layer). The first layer reads the words and maps them into an embedding space in a very similar way to the static models. Then the output of the first layer constitutes the input to the next layer, and so on (typically there are at least 12 layers). This design is considered superior to the design of static models, as it is thought to capture linguistic information at increasing levels of complexity and abstractness. The parameters of contextualized model are the aggregation of all the parameters in all the different layers.

Working with contextualized models is done differently than with previous models. Because of their large number of parameters, training contextualized models requires much more computational resources than previous model types. Most researchers do not have access to such large resources. As a result, many rely on pretrained models (i.e., models that were trained by others, and  made available online). 
  Current research often starts with searching for the most suitable pretrained model. One can supplement a pretrained model in a process called \textit{fine-tuning}, which is basically the continuation of the previous training on additional text. This step can be used to expose the model to text in a new domain (e.g., historical, technical) or train it on a new task, thus allowing the model to update based on a new type of information. Fine-tuning cannot undo (remove) the information a model has acquired during earlier stages from the corpus on which it was originally pretrained on. However, being aware of this problem allows us to adopt strategies that can take the original training corpus into account. Fine-tuning is considerably quicker than pretraining, and is therefore within the reach of most research labs. 

\subsection{Hyper-parameters}\label{sec:hyperparam}
Several factors influence the training process and  the way meaning is modeled from text. These factors are controlled by the  hyper-parameters of the model (not to be confused with the model's parameters described above). Table \ref{tab:hyperparameters} provides a comparison and overview of common hyper-parameters.
For static embeddings, hyper-parameters should be chosen at training time. Researchers rarely train contextualized models themselves, but it is important to be aware of the hyper-parameters chosen by the model providers. 

\begin{table}[!ht]
    \centering
    \caption{Hyper-parameters involved in models for meaning}\label{tab:hyperparameters}
    \begin{tabular}{|l|l|l|l|l|l|}
    \hline
        ~ & Window size & \# Epochs & Learning rate & Minimum Freq. & Down Sampling \\ \hline
        Count-based & + & ~ & ~ & ~ & ~ \\ \hline
        Static & + & + & + & + & + \\ \hline
        Contextualized & ~ & + & + & ~ & ~ \\ \hline
    \end{tabular}
\end{table}

The first hyper-parameter is the \textit{window size}, which determines how many context words a model considers to the left and right of the target word. As we derive meaning from usage, it is crucial to note that the number of words considered as context affects the type of linguistic information the model is exposed to. For example, with a window size of 2, which means two words before and two after the target word, a model will mainly be exposed to (and learn) syntactic information, and will filter out most of the semantic information that usually appears in a longer context. However, increasing the window size is only effective up to a certain point. Beyond that, adding more words will not add to the semantic information and can negatively influence the training time \citep{Levy-bullinaria-1998, bernier-colborne-drouin-2016-evaluation-distributional}.

Other hyper-parameters determine whether a model will converge, or in the geometric analogy, whether the word vectors will reach their desired semantic region. Convergence largely depends on whether the model was sufficiently trained. If it was trained on too few examples, then it is unlikely to converge, and the vector representations it produces are inaccurate and unreliable. \textit{Number of epochs} controls how long we train the model. A single epoch means that the model  completes a single pass over all the text, which is often insufficient. Naturally, if the corpus is small, more passes are required for the model to converge, and vice versa. In addition, \textit{learning rate} controls the size of the ``steps'' with which the model updates its vectors. A higher learning rate means that words reach their desired region in the semantic space quicker. However, words may overshoot their correct region, and be unable to make the small adjustments needed to reach their optimal location. On the other hand, a lower rate may mean that the training process takes an unreasonable amount of time to finish. Naturally, the learning rate and the number of epochs interact. A model may converge faster (requiring fewer epochs) if the learning rate is high, albeit with less fine-grained accuracy. In contrast, it takes more epochs for the model to converge if the learning rate is low.

Together, learning rate and number of epochs control the convergence of the model as a whole. However, if we are interested in whether certain words have converged or not, 
we must consider the frequency with which they appear in the text. High frequency words will be  updated more often than low frequency words, and are hence expected to converge faster, that is, they reach the desired region in semantic space faster than low frequency words \citep{dubossarsky2017outta}. \textit{Minimum frequency} in Table \ref{tab:hyperparameters} tells the model to ignore words below a certain frequency as their vectors are not expected to converge anyway.\footnote{It is advised to use a minimal threshold for computational reasons (e.g., memory load and training speed). However, practical advice would be to train a model with a minimum frequency of 10, and in subsequent analysis disregard frequencies higher than 100 or 200.} By contrast, \textit{down sampling} tells the model to occasionally ignore some high frequency words in proportion to their frequency (i.e., ultra-high frequency words will be ignored the most), because these words appear often enough to converge, and  updating them further will not generate better vectors, and will just extend the training time  unnecessarily.

\section{The computational study of semantic change}

To detect semantic change we employ a simple extension of the DH: If meaning is derived by analyzing the usage context of words, which are encoded in their vector representations, then changes to these representations must signify changes to their usage contexts and, by extension, their meanings. Put differently, after a model has converged, the vectors that represent the meaning of words occupy different areas in \textit{the semantic space} depending on their  meaning. Therefore, if a word has changed its meaning, so will its vector representation, and as a result the word will have moved to a different region of the space.

In order to measure these changes we need to compare a word at two or more time points. 
This can be done in one of two ways: 

First, we can look at a single word and compute the size of its displacement between two time points (the distance it traversed in the space) by comparing its vector representation in different time periods. 

Secondly, we can look at a word's position relative to its surrounding words (nearest neighbors). When words are compared in this way, it is because these neighbors are assumed to represent the category of meaning the word belongs to. In this case, we compute the distances of a word to its \textit{K}-nearest neighboring words, and evaluate how these distances change between two time periods. 
Usually a standard distance function such as cosine distance\footnote{Currently, the standard distance measure is the cosine distance between two vectors. However, given that these vectors can have extremely many dimensions, a very small change in one dimension could in theory signify a large change in meaning. Therefore, a future direction for research is to investigate and evaluate other distance measures that can account for small but significant changes. } is used to measure the distance, and we assume that larger distances between time periods correspond to greater changes in meaning. 

It is important to note that the method we use for detecting and analyzing semantic change is dependent on, and therefore limited by, the capacity of the computational method we choose to capture and represent meaning. In other words, if the model  we use is unable to capture meaning accurately, we will inevitably fail to detect semantic change. Likewise, if a model captures only certain aspects of meaning, for example, syntactic information, we will fail to detect change in other aspects. 

\subsection{Computationally detecting semantic change} 
\label{ChoiceOfComputataionModel}
The task of semantic change detection is defined according to our choice of computational model of meaning (see Section 3). However, all models share a few steps: First the historical data is divided into non-overlapping \textit{time bins} (see more in 3.2). Then, vector representations are created for every \textit{target word} individually for each time bin. Typically, we create vectors for all words in the lexicon, but a subset is also possible. Finally, the vectors of each word corresponding to different time points are compared and their change is assessed. 
Below we focus on unique steps and the relevant considerations for different models. 

\bigskip\noindent\textbf{Count-based models}. Change detection using count-based vectors is dependent on whether the vectors are word-level vectors or token vectors. For word-level vectors, the situation is straightforward. 
Count-based models generate explicit vector representations for the target words. Each cell in the vector corresponds to a specific word that co-occurs with the target word (or its context words). The dimensions (elements in the vectors) are known and can be fixed across time, for example, as the union of all words that occur in any time periods sorted in alphabetic order. Therefore,  a distance function can directly be used to measure how much the vectors corresponding to a word have changed between two time points.  Because of the explicit nature of the vectors, it is possible to examine which words co-occur more or less frequently with the target word and thus   investigate the nature of any change. For example, the model may detect that the word \textit{Apple} now more frequently co-occurs with technical products than with fruit and other edibles.  

\bigskip\noindent\textbf{Static models}. Predictive static models train vector spaces on text for each time bin independently, resulting in separate vectors spaces for each bin. 
Because the dimensions of the vector space are neither explicit nor known, vectors from independent spaces cannot be directly compared. In the room analogy, one can think of this as animals being located close together, but in different corners of the room at different points in time. 
Before we can compare vectors from different time periods, we must thus ensure that the vectors belong to the same space.

There are three ways in which this problem can be alleviated: (a) Aligning pairs of vector spaces prior to comparison (i.e., rotating the spaces such that their basis axes  match) \citep{hamilton2016diachronic}. (b) Using       temporal referencing \citep{dubossarsky2019timeout} that uses the original corpus without dividing it into time bins and adds temporal information to the target words as special tags. This means creating several versions of a word (cell\textunderscore{1910}, cell\textunderscore{1920}, etc.) and putting all words in a single vector space. (c) Using incremental training \citep{kim2014temporal} where the vector space of each following time bin starts with (i.e., is initialized with) the vector space of the previous time bin. All three methods ensure that direct comparison between vectors from different time bins is possible using a distance function based on vector arithmetic. 

If one uses the method of nearest neighbors, individually trained vector spaces can be used for change detection without prior alignment. In this case, the word is compared only to its neighbors from the same vector space, ensuring that the vectors already share the space and can thus be directly compared.

\bigskip\noindent\textbf{Contextualized models}. Contextualized models create token vectors, that is, each individual occurrence of a word is represented by its own vector. The comparison of vectors from different time periods is trivial: all vectors share the same semantic space as they are derived from the same model. 
Change detection then follows in one of three ways:\footnote{Though it has not been done regularly, a 
 similar methodology can be used for count-based models.}
\begin{enumerate}
    \item Token vectors are averaged across all occurrences (creating a single type vector for each word per time period) and comparison is done by using a distance function, as with the static model. The benefits compared to static models stem from the model's training on a much larger textual corpus, resulting in significantly richer information that is expected to carry over to the averaged vector.
    \item Token vectors are first clustered according to similarity in meaning, and then compared across time periods. Then we can either follow the above procedure and average the token vectors of all occurrences that belong to the same sense, or use a distance function (e.g., Jensen-Shannon) that measures the changing sense distributions over time \citep{giulianelli-etal-2020-analysing}. The benefits of the approach are clear: we can derive sense information from the token representations. However, the change detection problem becomes significantly more complex: instead of comparing a single vector per time period, we need to compare several for each time period. 

\end{enumerate}

\noindent Using contextualized models does not usually require training, as models are generally pretrained but can optionally be fine-tuned on a historical corpus.\footnote{We will use ``historical corpus'' as an example here, but this corpus could be exchanged for any other domain-specific corpus, for example containing medical texts or literature. } However, unlike count-based and static models that  generate their representations solely from the (historical) corpus they are trained on, contextualized models are trained on a huge amount of  ``all purpose'' text before being applied to historical text. This causes a major concern as vectors generated from historical text may contain information picked up during pretraining from a different domain, leading to meaning representations that cannot be attributed to historical information alone.\footnote{Although it is possible in theory to pretrain a contextualized model completely on historical text and thus eliminate this problem, this would require a huge amount of data that is typically not available for historical texts.} Despite this attribution problem, it is advantageous to use contextual models precisely because their pretraining on massive amounts of data leads to meaning representations of higher quality. As most words are rather stable in meaning historically (especially over short time periods), contextualized models provide more accurate representations than models that are trained only on historical data whose size is orders of magnitude smaller. A smaller corpus will have difficulty producing accurate vectors even for stable words. The problem of information prevalence  in pretrained models should not prevent their use, but users should be aware of this issue and place a higher emphasis on testing and evaluation of a detected change.

\bigskip\noindent\textbf{Choice of model}. So far, we have seen strong evidence that count-based models for semantic change are outperformed by predictive models. Research on predictive models has also shown some indication, although no conclusive evidence, that contextual methods outperform static methods in the task of computational modeling of semantic change. Therefore, at present the choice between the two methods can be made on the basis of the research question rather than quality indicators and  the availability of pretrained models and historical corpora. We advise that the community should not focus exclusively on one type of meaning representation as it is likely that different research formulations (see section \ref{subsec:researchformulations}) will need different meaning representations, in particular those that go beyond the distributional hypothesis.

\subsection{CUI: The corpus under investigation}
Like most data-intensive research fields, computational modeling of semantic change is dependent on the quality of the data it processes. We call this data the corpus under investigation (CUI), where the word \textit{corpus} refers to either a collection of texts (i.e. a corpus in the linguistic sense) or a  (thematic) dataset. 

CUIs differ significantly depending on the source and the types of semantic change recorded in them. For example, the potential changes that can be found in a digitized corpus of newspapers spanning 200 years are different from those found in social media texts spanning 5 to 10 years. Likewise, analyzing standard language may generate different vectors for the same words when compared to more specialized domains that contain text on different topics or in different genres or registers (e.g., medical, financial, or legal).  

Therefore, a prerequisite for semantic change analysis is that the changes we are looking for can potentially appear in the CUI (and in large enough quantity to be detectable) both in terms of genre and time periods. It is also important that the CUI has temporal information (relating to day, month, year, or decade of writing).

A CUI is commonly divided into non-overlapping time bins to facilitate its historical analysis. Alternatively, different corpora distributed over time are used, like the British National Corpus (BNC), or a modern newspaper collection.  In all analyses, we assume that changes only exist \textit{between} time bins, and that no (or negligible) change exists \textit{within} an individual time bin. One can control the binning granularity, creating CUIs with finer or coarser time resolution, to better suit the nature of the semantic change being studied (e.g., ephemeral change requires shorter time bins). However, more binning means that the amount of text per bin is  reduced, and depending on the size of the original corpus, there is a trade-off between high granularity and a sufficient amount of text per time bin to draw reliable conclusions.\footnote{Binning has particularly adverse effects for static models that require each word to have many repetitions during training.    The decreased frequency of words per time bin reduces the quality of word representations, and consequently reduces  the ability of the model to reliably detect genuine changes in word meaning. See also 4.1.}

Other characteristics to consider are spelling variations, either due to mistakes, lack of canonical spelling (especially in older texts and social media) or OCR errors. A high amount of nonstandard spelling reduces the number of instances of a given word, as they will be split over multiple different word forms. The preprocessing steps are also relevant, for example, lemmatization or syntactic information. A non-lemmatized text would result in a different vector representation for each inflection of a word, thus complicating its analysis. This also results in fewer occurrences of a word type and thus leads to lower quality (static) vector spaces (unless the CUI is huge). However, for contextualized methods, lemmatization is not necessary. By contrast, syntactically parsed text would enable us to easily discern that \textit{Apple} changed its meaning by adding a meaning related to a proper noun (the company), leaving its fruit sense unchanged.  

\subsection{Choosing the CUI}

Our research aims and boundaries directly affect the choice of CUI and the corresponding computational method. For example, a model that cannot handle large volumes of text cannot be used for a CUI with billions of words. Or, if we aim to find semantic change in medical terminology, we need a method that can create meaning representations for multi-word expressions.  
 We should therefore ensure that (a) the CUI contains answers to our research questions and (b) that our models can accurately represent meaning needed to answer our questions based on that particular CUI.

Importantly, when comparing corpora over time, we should be aware of the possibility that other confounding variables may have changed. Often, due to lack of good, long-spanning datasets, corpora from different mediums are compared, leading to a simultaneous investigation of both semantic change and, for example, genre change. Despite the current availability of historical corpora being less than optimal, we can make use of the available data provided we verify the results using proper evaluation techniques (see  Section \ref{sec:eval}).

When it comes to finding answers to the general question  \textit{Have our words changed their meaning?} we need general and broad corpora. However, even the most comprehensive, balanced, or curated CUI constitutes only a sample of language usage and, since it cannot be said to be a random sample, it cannot represent the language as a whole (for a longer discussion, see \cite{Koplenig2016} and \cite{kutuzov2020-phd}). As a result, we can miss words that have undergone semantic change, get a skewed picture of these changes, or find corpora-specific changes that do not generalize. Ultimately, it is difficult to know whether the results (or lack thereof) represent how language truly behaves, or whether they stem from a bias in the CUI. The same holds for the interaction between the CUI and the computational methods we choose \citep{dubossarsky2017outta}. One should always exercise caution when reporting findings (for example, listing a set of words that are historically stable or deriving laws of semantic change). To partially remedy this, and to get more general results, it is advisable to repeat the analyses multiple times under different conditions. In others words, it is important to use different CUIs and computational models, and to conduct statistical tests to evaluate how likely these changes are relative to random effects. Here, pretrained models that make use of huge background corpora are better suited, specially if their results are compared to the results of models trained on a known CUI.

\section{Evaluation }\label{sec:eval}

As is the case in most data-driven and experimental research, evaluation is very important. We want to make sure that the information we are capturing is \textit{correct}, and also to some extent \textit{complete}. The standard of evaluation in lexical semantic change research has fluctuated considerably so far. Before the introduction of standard datasets and evaluation tasks, researchers were left to evaluate as they saw fit. 

The first standard evaluation task was launched in 2020: SemEval-2020 Task 1 on unsupervised lexical semantic change detection \citep{schlechtweg-etal-2020-semeval}.  Since then, we have seen several more shared tasks and a move towards a standard of evaluation. To date, there are evaluation datasets for German, English, Swedish, Latin, Russian, Italian, Spanish, and Norwegian \citep{schlechtweg-etal-2020-semeval,diacrita_evalita2020,  kutuzov2021rushifteval, kutuzov-etal-2022-nordiachange, Zamora2022lscd}. 

Because our models of meaning rely on vector representations, we must begin by evaluating the quality of a vector space. In doing this, we indirectly evaluate the accurate meaning representation of our words (especially for static embedding representations). With pretrained contextualized models, the quality of the vector space is often assumed to be high due to its ubiquitous use in other domains. Nonetheless, we encourage proper testing, for example, by manually investigating a few words and their relations.

Depending on  what question we set out to answer (and recall the problem formulations in section \ref{subsec:researchformulations}), we need to perform different kinds of evaluation. If our aim is to find what is present in a CUI, we need to ground the results in the CUI. Does the CUI support the finding that a word has changed its meaning in the way  the model proposes? If, however, what we are looking for relates to language in general, we can evaluate our results using dictionaries or by asking annotators about their general knowledge of the semantic change of a word. Often, the examples used in evaluation  are  novel words or words that have experienced large semantic changes. For example, novel words from a technical domain like \textit{Internet}, \textit{blog}, or \textit{wifi} are often used to control whether the model can find what they mean and when these meanings appeared.  Examples of words with large semantic changes that are used for evaluation include  \textit{mouse}, which adds a technical meaning; \textit{rock}, which adds a music sense; and \textit{Apple}, which adds the meaning of a company or product to its existing senses. In clear-cut cases like these, evaluation can be done by comparing the outcome of a model to general world knowledge, found either in dictionaries or in knowledge bases like Wikipedia. In \citet{tahmasebi2021survey}, the word \textit{computer} is used to evaluate three different CUIs from the Google N-gram corpus. The frequency of the word "computer" (\textit{Rechner} in German)
rises above 0.0001\% in 1934 for  German, 1943 for  American English, and 1953 for  British English. This example shows that a method evaluated on the
latter dataset would be penalized by 20 years compared to one evaluated on the German dataset. Therefore, even in generic cases of semantic change, it is important to ground the results in the relevant CUI to know the true performance of the method's ability to detect change. 
  
When it comes to evaluating unknown words, or detecting change for all words in the vocabulary, there are still two key questions we need to be able to answer that nicely summarize the semantic change task itself, namely, what a word means and when two meanings are the same.  

\subsection{A thought experiment}

To illustrate the difficulty in modeling both meaning and meaning change, let us conduct a thought experiment. We have a corpus, M,  of decent size, let us say 200 million tokens. If we model the meaning of a word given the whole corpus, we will get a meaning representation \textit{w(M)} for a word \textit{w}.   \textit{w(M)} can be either a single representation, as in type-based vector representations, or a set of representations as in contextual models. 

\paragraph{What does a word mean?} To answer this first question,  we must ensure that the function we have chosen to map a word \textit{w} to a representation \textit{w(M)} is appropriate. Put differently, we need to know that \textit{w(M)} provides an accurate representation of \textit{w} as it is evident in the whole corpus; it should not miss out on certain important parts, nor misrepresent \textit{w}. Given \textit{w(M)}, we should be able to understand what \textit{w} means in a way that is accurate according to all the information that is stored in our corpus \textit{M}. Today, these representations are often evaluated in one of two ways. The second seems more prominent, but is currently limited to modern meaning. 
\begin{enumerate}
\item A few sample words are chosen and their closest neighbors are investigated by manual inspection. Do the \textit{K}-nearest neighbors correspond to the expected result? To give an example, does the word \textit{rock} have neighbors that relate only to \textit{natural stone} in historical text and to both \textit{stone} and \textit{music} in modern text? 
\item The vector space is evaluated by measuring how well it preserves semantic similarity according to a standard word similarity dataset. Examples of these datasets include WordSim-353 \citep{Finkelstein:2001} and SimLex-999 \citep{hill2015simlex}. Each of these datasets contains a set of words that have been rated for semantic similarity, or semantic relatedness, by human annotators.
\end{enumerate}

\paragraph{When are two meanings the same?}
The second question we need to answer is when two meanings are the same. Imagine that we split the corpus $M$ into two halves, $M_1$ and $M_2$, each containing 100 million tokens.\footnote{This is still a reasonable size to assume that our models are providing robust results and not random noise. } From each of these halves, we get one representation for the word \textit{w}. Let us call them $w(M_1$) and $w(M_2$).  Because the three corpora, $M$, $M_1$ and $M$$_2$ all have different tokens, the three representations for the same word, $w(M)$, $w(M_1$) and $w(M_2$), will all be different. In fact, each time we make a new random split resulting in new subcorpora $M_1$ and $M_2$, the representations will be different.\footnote{This is true unless we can assume that the size of $M, M$$_1$ and $M$$_2$ approaches infinity. However, because few of our datasets have the property of having infinitely many tokens, we can safely assume that the corpus M is of finite size, and hence  both $M$$_1$ and $M$$_2$ are of finite size.} 
The differences between $w(M)$, $w(M$$_1$) and $w(M$$_2$) are due to random variations, or variance in the tokens underlying the modeling. However, because there is no 'true' semantic change, as we have just randomly split our dataset in two, we want to have a measure that can detect that $w(M), w(M$$_1$) and $w(M$$_2$) are in fact similar enough to represent the same word $w$. 

Now let us assume that the corpora $M, M$$_1$ and $M$$_2$ are  from three different time periods $t_0, t_1, t_2$, each with 100 million tokens. We will still end up with three different representations of the meaning of word $w$ namely $ w(M,$ $t_0)$, $w(M_1, t_1)$ and $w(M_2, t_2)$. But, unlike the case above where the differences  were insignificant, the differences here may be due to gradual semantic change, and should  be detected as such by the change detection algorithm. 

In conclusion, we want to be able to distinguish between  true cases of semantic change and those that stem from random  variation in the underlying dataset. We recognize that semantic change scores are reliable only if they are larger than the scores of random variations \citep{dubossarsky2017outta, dubossarsky2019timeout}.

\subsection{Evaluation strategies}
Generally speaking, semantic change is evaluated using one of three strategies: (1) evaluate on a testset previously evaluated by hand; (2) evaluate the outcome of an experiment; and (3) evaluate using control settings. We will briefly discuss these different evaluation strategies and their pros and cons. 

\paragraph{A priori determined testsets.} In this category of testing we determine what to expect for a smaller selection of words before we start  the experiments. Typically, these words are tied to the corpus that we are using, that is, the CUI. Sometimes, however, these are lists of words with an attached change score that are produced without regard to any dataset,  for example,  the one provided by \cite{Gulordava11}. 
Examples where the testset is tied to the CUI are the SemEval and follow-up testsets, where a set of words have been manually determined to belong to the categories of Change or Stable in the corresponding CUIs. In evaluation tasks like SemEval-2020 Task 1, this is called semantic change \textit{detection}. 

The testing of both changing and stable words is important. If we test a model's ability to determine change only on words that have experienced change, we will not be able to detect if the model classifies all words as changing words. Hence, a model's ability to differentiate between changed and stable words is important. 

Creating these evaluation datasets requires extensive manual effort. For that reason, they are often restricted in various ways, for example, in the number of time points that are used for evaluation, or the number of words for which change scores are provided. However, by putting in the effort a priori, we can compare any results, regardless of how they were created. Every time new algorithms or modeling paradigms appear, we can retrain our models, run the experiments and evaluate them with respect to known testsets and compare our results to previous results under the same conditions. 

\paragraph{Evaluating a posteriori.} Instead of evaluating with respect to a preexisting testset, we can choose to evaluate the \textit{outcome of our models} directly. This evaluation framework is important when we target novel task formulations or datasets for which there are no a priori available testsets. 
The downsides of this kind of evaluation are, however, plentiful. Firstly, the conditions under which we evaluate are different each time. The majority of the work on semantic change was evaluated in this fashion in the past. Each researcher defined their own evaluation criteria, size of evaluation set, and often cherry picked the examples for illustration, thus providing results that cannot be compared. 
Another clear downside is that we get different results each time we make changes to our methods, use different preprocessing, or make changes to the hyper-parameters. These results must then be re-evaluated. In the worst case, if there is no overlap between the current and the previous set of results, all the manual effort needs to be repeated. In the LSCDiscovery shared task for Spanish \citep{Zamora2022lscd}, this is called semantic change \textit{discovery}.

The upsides of this method are nonetheless substantial. This evaluation framework allows us to go beyond predetermined sets of words and datasets to find out how our algorithms perform 'in the wild'. There is a clear benefit to using this framework, and we suggest it is used as an \textit{important complement} to the a priori evaluation framework. First, we can find out how our method performs on a known testset compared to other methods. We may possibly employ these preliminary tests to  develop or tune the model. Then, we can run it on a new dataset or in a new context and evaluate the outcome of the method. This combination reduces the amount of effort required for evaluating while still providing valuable insights into the performance of the model. 

\paragraph{Evaluating using control settings.}
In addition to evaluating how our methods are able to detect semantic change on the level of individual words, we can evaluate what our methods find on the level of the vocabulary as a whole. Providing manual evaluation data on the level of the vocabulary would be extremely costly in terms of both time and money. We therefore need to devise other strategies for controlling our methods. We can do this either by using synthetic evaluation data where we simulate change on a larger scale, as in the work of \cite{shoemark2019room} and \citet{dubossarsky2019timeout}, or by using other control criteria. One example of a control criterion is \textit{shuffling}, which exposes inherent properties of the corpus as a whole and allows us to test whether our methods are detecting semantic change or other underlying change properties. If every time period contains data from every other time period, then we should not expect our methods to detect any semantic change. But if the methods do detect  semantic change after shuffling of the texts, then we can assume that they are, in fact, modeling something other than semantic change, possibly related to fluctuations in the amount of data over time.   

In general, the majority of the corpora we use for semantic change detection naturally grow in size over time. For example, the majority of our newspaper corpora have an increasing number of articles as well as longer articles over time. The question that we need to answer is the following: Are we detecting semantic change or the fact that our datasets are larger? We can control for this increasing data size by, for example,  using a fixed amount of text per time period.

These kind of control criteria allow us to zoom out from the individual details and obtain a high-level overview of the model in general. That our models can detect 'true' semantic change in individual instances is beside the point -- even a broken watch is correct twice per day. It is the general behavior over the whole vocabulary and time period that is important. Our methods should not be capturing change when there is no change to be found.

\section{Summary}
Computational modeling of semantic change can be used as a powerful tool to study individual words, whole languages, and the interaction between languages. The models can be used to study the process of change, how changes spread across languages, and how languages affect each other.  Computational models of semantic change can also be used to study changing perceptions and change in cultural and societal aspects of our lives as evidenced in text.   With these models, we have the possibility of studying changing phenomena that live in our texts, regardless of whether they are describing our language or our lives beyond the text. Thus far, however, models for semantic change have not been used to their full potential and there is much research to be done  in terms of both development within semantic change and its application to other fields. 

In this chapter, we have discussed different methods for modeling meaning based on given texts, as well as modeling meaning change. We have described the latest methods in terms of modeling, both static and contextual embedding methods, as well as their pros and cons with respect to the study of semantic change. We have provided some insight into the interaction of data sources and semantic change detection and have discussed important aspects relating to evaluation of the results.  

This chapter in the handbook should enable researchers with a general knowledge of natural language processing, (computational) linguistics, or computer science to begin modeling semantic change. They will have the tools to choose an appropriate model and appropriate data to correspond to the research question, and they will be able to perform appropriate evaluation of the results. 

\section*{Acknowledgement}

This work has been funded in part by
the project \textit{Towards Computational Lexical Semantic Change Detection }supported by the Swedish Research Council (2019–2022; contract 2018-01184), by the research program \textit{Change is Key!} supported by Riksbankens Jubileumsfond (under reference number M21-0021) and by \textit{Cassandra} supported by the Marcus and Amalia Wallenberg Foundation (MAW 2020.0060). We would also like to acknowledge Antske Fokkens for inspiring the room analogy in Section \ref{sec:geom_interp}.

\bibliographystyle{abbrvnat}
\bibliography{sample, newbibliography, localbibliography}

\begin{thebibliography}{33}
\providecommand{\natexlab}[1]{#1}
\providecommand{\url}[1]{\texttt{#1}}
\expandafter\ifx\csname urlstyle\endcsname\relax
  \providecommand{\doi}[1]{doi: #1}\else
  \providecommand{\doi}{doi: \begingroup \urlstyle{rm}\Url}\fi

\bibitem[Anttila(1989)]{anttila1989historical}
R.~Anttila.
\newblock \emph{Historical and Comparative Linguistics}, volume~6.
\newblock John Benjamins Publishing, 1989.

\bibitem[Basile et~al.(2020)Basile, Caputo, Caselli, Cassotti, and
  Varvara]{diacrita_evalita2020}
P.~Basile, A.~Caputo, T.~Caselli, P.~Cassotti, and R.~Varvara.
\newblock {Overview of the EVALITA 2020 Diachronic Lexical Semantics
  (DIACR-Ita) Task}.
\newblock In V.~Basile, D.~Croce, M.~Di~Maro, and L.~C. Passaro, editors,
  \emph{{Proceedings of the 7th evaluation campaign of Natural Language
  Processing and Speech tools for Italian (EVALITA 2020)}}, Online, 2020.
  CEUR.org.

\bibitem[Bernier-Colborne and
  Drouin(2016)]{bernier-colborne-drouin-2016-evaluation-distributional}
G.~Bernier-Colborne and P.~Drouin.
\newblock Evaluation of distributional semantic models: A holistic approach.
\newblock In \emph{Proceedings of the 5th International Workshop on
  Computational Terminology (Computerm2016)}, pages 52--61, Osaka, Japan, Dec.
  2016. The COLING 2016 Organizing Committee.
\newblock URL \url{https://aclanthology.org/W16-4707}.

\bibitem[Bizzoni et~al.(2020)Bizzoni, Degaetano-Ortlieb, Fankhauser, and
  Teich]{BizzoniDegaetanoOrtliebFankhauseretal-2020}
Y.~Bizzoni, S.~Degaetano-Ortlieb, P.~Fankhauser, and E.~Teich.
\newblock Linguistic variation and change in 250 years of {E}nglish scientific
  writing: A data-driven approach.
\newblock 3\penalty0 (73), 2020.
\newblock ISSN 2624-8212.
\newblock \doi{10.3389/frai.2020.00073}.
\newblock URL \url{http://nbn-resolving.de/urn:nbn:de:bsz:mh39-100889}.

\bibitem[Brunila and LaViolette(2022)]{Brunila2022}
M.~Brunila and J.~LaViolette.
\newblock {What company do words keep? Revisiting the distributional semantics
  of J.R. Firth and Zellig Harris}.
\newblock \emph{NAACL 2022 Conference of the North American Chapter of the
  Association for Computational Linguistics: Human Language Technologies,
  Proceedings of the Conference}, pages 4403--4417, 2022.
\newblock \doi{10.18653/v1/2022.naacl-main.327}.
\newblock URL \url{https://aclanthology.org/2022.naacl-main.327.pdf}.

\bibitem[Dubossarsky et~al.(2017)Dubossarsky, Weinshall, and
  Grossman]{dubossarsky2017outta}
H.~Dubossarsky, D.~Weinshall, and E.~Grossman.
\newblock Outta control: Laws of semantic change and inherent biases in word
  representation models.
\newblock In \emph{Proceedings of the 2017 Conference on Empirical Methods in
  Natural Language Processing}, pages 1136--1145, 2017.

\bibitem[Dubossarsky et~al.(2019)Dubossarsky, Hengchen, Tahmasebi, and
  Schlechtweg]{dubossarsky2019timeout}
H.~Dubossarsky, S.~Hengchen, N.~Tahmasebi, and D.~Schlechtweg.
\newblock Time-out: Temporal referencing for robust modeling of lexical
  semantic change.
\newblock In \emph{Proceedings of the 57th Annual Meeting of the Association
  for Computational Linguistics (Volume 1: Long Papers)}, Florence, Italy,
  2019. Association for Computational Linguistics.

\bibitem[Finkelstein et~al.(2001)Finkelstein, Gabrilovich, Matias, Rivlin,
  Solan, Wolfman, and Ruppin]{Finkelstein:2001}
L.~Finkelstein, E.~Gabrilovich, Y.~Matias, E.~Rivlin, Z.~Solan, G.~Wolfman, and
  E.~Ruppin.
\newblock Placing search in context: The concept revisited.
\newblock In \emph{Proceedings of the 10th International Conference on World
  Wide Web}, WWW '01, pages 406--414, New York, NY, USA, 2001. ACM.

\bibitem[Frermann and Lapata(2016)]{frermann2016bayesian}
L.~Frermann and M.~Lapata.
\newblock A {B}ayesian model of diachronic meaning change.
\newblock \emph{Transactions of the Association for Computational Linguistics},
  4:\penalty0 31--45, 2016.

\bibitem[Giulianelli et~al.(2020)Giulianelli, Del~Tredici, and
  Fern{\'a}ndez]{giulianelli-etal-2020-analysing}
M.~Giulianelli, M.~Del~Tredici, and R.~Fern{\'a}ndez.
\newblock Analysing lexical semantic change with contextualised word
  representations.
\newblock In \emph{Proceedings of the 58th Annual Meeting of the Association
  for Computational Linguistics}, pages 3960--3973, Online, July 2020.
  Association for Computational Linguistics.

\bibitem[Gulordava and Baroni(2011)]{Gulordava11}
K.~Gulordava and M.~Baroni.
\newblock A distributional similarity approach to the detection of semantic
  change in the {Google Books Ngram} corpus.
\newblock In \emph{{Proceedings of the Workshop on Geometrical Models of
  Natural Language Semantics}}, pages 67--71, Stroudsburg, PA, USA, 2011.

\bibitem[Hamilton et~al.(2016)Hamilton, Leskovec, and
  Jurafsky]{hamilton2016diachronic}
W.~L. Hamilton, J.~Leskovec, and D.~Jurafsky.
\newblock Diachronic word embeddings reveal statistical laws of semantic
  change.
\newblock In \emph{Proceedings of the 54th Annual Meeting of the Association
  for Computational Linguistics}, volume~1, pages 1489--1501, 2016.

\bibitem[Hill et~al.(2015)Hill, Reichart, and Korhonen]{hill2015simlex}
F.~Hill, R.~Reichart, and A.~Korhonen.
\newblock Simlex-999: Evaluating semantic models with (genuine) similarity
  estimation.
\newblock \emph{Computational Linguistics}, 41\penalty0 (4):\penalty0 665--695,
  2015.

\bibitem[Kim et~al.(2014)Kim, Chiu, Hanaki, Hegde, and Petrov]{kim2014temporal}
Y.~Kim, Y.-I. Chiu, K.~Hanaki, D.~Hegde, and S.~Petrov.
\newblock Temporal analysis of language through neural language models.
\newblock \emph{ACL 2014}, page~61, 2014.

\bibitem[Koplenig(2016)]{Koplenig2016}
A.~Koplenig.
\newblock \emph{Analyzing lexical change in diachronic corpora}.
\newblock PhD thesis, Mannheim, 2016.
\newblock URL \url{http://nbn-resolving.de/urn:nbn:de:bsz:mh39-48905}.

\bibitem[Kutuzov(2020)]{kutuzov2020-phd}
A.~Kutuzov.
\newblock \emph{Distributional word embeddings in modeling diachronic semantic
  change}.
\newblock PhD thesis, University of Oslo, 2020.

\bibitem[Kutuzov et~al.(2021)Kutuzov, Pivovarova,
  et~al.]{kutuzov2021rushifteval}
A.~Kutuzov, L.~Pivovarova, et~al.
\newblock Rushifteval: a shared task on semantic shift detection for {R}ussian.
\newblock In \emph{Computational Linguistics and Intellectual Technologies:
  Papers from the Annual Conference Dialogue.} Redkollegija sbornika, 2021.

\bibitem[Kutuzov et~al.(2022)Kutuzov, Touileb, M{\ae}hlum, Enstad, and
  Wittemann]{kutuzov-etal-2022-nordiachange}
A.~Kutuzov, S.~Touileb, P.~M{\ae}hlum, T.~Enstad, and A.~Wittemann.
\newblock {N}or{D}ia{C}hange: Diachronic semantic change dataset for
  {N}orwegian.
\newblock In \emph{Proceedings of the Thirteenth Language Resources and
  Evaluation Conference}, pages 2563--2572, Marseille, France, June 2022.
  European Language Resources Association.
\newblock URL \url{https://aclanthology.org/2022.lrec-1.274}.

\bibitem[Lau et~al.(2014)Lau, Cook, McCarthy, Gella, and
  Baldwin]{lau2014learning}
J.~H. Lau, P.~Cook, D.~McCarthy, S.~Gella, and T.~Baldwin.
\newblock Learning word sense distributions, detecting unattested senses and
  identifying novel senses using topic models.
\newblock In \emph{Proceedings of the 52nd Annual Meeting of the Association
  for Computational Linguistics (Volume 1: Long Papers)}, volume~1, pages
  259--270, 2014.

\bibitem[Lehrer(1985)]{LEHRER+1985+283+296}
A.~Lehrer.
\newblock The influence of semantic fields on semantic change.
\newblock In J.~Fisiak, editor, \emph{Historical Semantics - Historical
  Word-Formation}, pages 283--296. Berlin: De Gruyter Mouton, 1985.
\newblock \doi{10.1515/9783110850178.283}.

\bibitem[Levy and Bullinaria(1998)]{Levy-bullinaria-1998}
J.~Levy and J.~Bullinaria.
\newblock Explorations in the derivation of semantic representations from word
  co-occurrence statistics.
\newblock \emph{South Pacific Journal of Psychology}, 10\penalty0 (1):\penalty0
  99--111, 1998.
\newblock \doi{10.1017/S0257543400001061}.

\bibitem[Marjanen et~al.(2019)Marjanen, Pivovarova, Zosa, and
  Kurunm{\"a}ki]{marjanen2019clustering}
J.~Marjanen, L.~Pivovarova, E.~Zosa, and J.~Kurunm{\"a}ki.
\newblock Clustering ideological terms in historical newspaper data with
  diachronic word embeddings.
\newblock In \emph{5th International Workshop on Computational History,
  HistoInformatics 2019}. CEUR-WS, 2019.

\bibitem[Mitra et~al.(2015)Mitra, Mitra, Maity, Riedl, Biemann, Goyal, and
  Mukherjee]{mitra2015automatic}
S.~Mitra, R.~Mitra, S.~K. Maity, M.~Riedl, C.~Biemann, P.~Goyal, and
  A.~Mukherjee.
\newblock An automatic approach to identify word sense changes in text media
  across timescales.
\newblock \emph{Natural Language Engineering}, 21\penalty0 (5):\penalty0
  773--798, 2015.

\bibitem[Schlechtweg et~al.(2020)Schlechtweg, McGillivray, Hengchen,
  Dubossarsky, and Tahmasebi]{schlechtweg-etal-2020-semeval}
D.~Schlechtweg, B.~McGillivray, S.~Hengchen, H.~Dubossarsky, and N.~Tahmasebi.
\newblock {S}em{E}val-2020 task 1: {U}nsupervised {L}exical {S}emantic {C}hange
  {D}etection.
\newblock In \emph{{Proceedings of the 14th International Workshop on Semantic
  Evaluation}}, Barcelona, Spain, 2020. Association for Computational
  Linguistics.

\bibitem[Sch\"{u}tze(1998)]{Schutze1998}
H.~Sch\"{u}tze.
\newblock Automatic word sense discrimination.
\newblock \emph{Computational Linguistics}, 24\penalty0 (1):\penalty0 97--123,
  Mar. 1998.

\bibitem[Shoemark et~al.(2019)Shoemark, Liza, Nguyen, Hale, and
  McGillivray]{shoemark2019room}
P.~Shoemark, F.~F. Liza, D.~Nguyen, S.~Hale, and B.~McGillivray.
\newblock Room to {G}lo: A systematic comparison of semantic change detection
  approaches with word embeddings.
\newblock In \emph{Proceedings of the 2019 Conference on Empirical Methods in
  Natural Language Processing and the 9th International Joint Conference on
  Natural Language Processing (EMNLP-IJCNLP)}, pages 66--76, Hong Kong, China,
  Nov. 2019. Association for Computational Linguistics.
\newblock \doi{10.18653/v1/D19-1007}.
\newblock URL \url{https://www.aclweb.org/anthology/D19-1007}.

\bibitem[Tahmasebi et~al.()Tahmasebi, Kutuzov, Giulianelli, and
  Dubossarsky]{Tahmasebi2024}
N.~Tahmasebi, A.~Kutuzov, M.~Giulianelli, and H.~Dubossarsky.
\newblock Computational modelling of semantic change.
\newblock In E.~Ledgeway, Adam~Aldridge, A.~Breitbarth, K.\`E. Kiss, J.~Salmons,
  and A.~Simonenko, editors, \emph{Wiley Blackwell Companion to Diachronic
  Linguistics (DiaCom)}. Wiley Blackwell. \textit{In preparation}.

\bibitem[Tahmasebi et~al.(2021)Tahmasebi, Borin, and
  Jatowt]{tahmasebi2021survey}
N.~Tahmasebi, L.~Borin, and A.~Jatowt.
\newblock Survey of computational approaches to lexical semantic change
  detection.
\newblock In N.~Tahmasebi, L.~Borin, A.~Jatowt, Y.~Xu, and S.~Hengchen,
  editors, \emph{Computational Approaches to Semantic Change}, Language
  Variation, chapter~1. Language Science Press, Berlin, 2021.

\bibitem[Tahmasebi(2013)]{tahmasebi2013models}
N.~N. Tahmasebi.
\newblock \emph{Models and Algorithms for Automatic Detection of Language
  Evolution}.
\newblock PhD thesis, Gottfried Wilhelm Leibniz Universit\"{a}t Hannover, november
  2013.
\newblock URL
  \url{http://edok01.tib.uni-hannover.de/edoks/e01dh13/771705034.pdf}.

\bibitem[Tripodi et~al.(2019)Tripodi, Warglien, Levis~Sullam, and
  Paci]{tripodi-etal-2019-tracing}
R.~Tripodi, M.~Warglien, S.~Levis~Sullam, and D.~Paci.
\newblock Tracing antisemitic language through diachronic embedding
  projections: {F}rance 1789-1914.
\newblock In \emph{Proceedings of the 1st International Workshop on
  Computational Approaches to Historical Language Change}, pages 115--125,
  Florence, Italy, Aug. 2019. Association for Computational Linguistics.
\newblock \doi{10.18653/v1/W19-4715}.
\newblock URL \url{https://www.aclweb.org/anthology/W19-4715}.

\bibitem[Vylomova et~al.(2019)Vylomova, Murphy, and
  Haslam]{vylomova-etal-2019-evaluation}
E.~Vylomova, S.~Murphy, and N.~Haslam.
\newblock Evaluation of semantic change of harm-related concepts in psychology.
\newblock In \emph{Proceedings of the 1st International Workshop on
  Computational Approaches to Historical Language Change}, pages 29--34,
  Florence, Italy, Aug. 2019. Association for Computational Linguistics.
\newblock \doi{10.18653/v1/W19-4704}.
\newblock URL \url{https://www.aclweb.org/anthology/W19-4704}.

\bibitem[Xu and Kemp(2015)]{xu2015computational}
Y.~Xu and C.~Kemp.
\newblock A computational evaluation of two laws of semantic change.
\newblock In \emph{CogSci}, 2015.

\bibitem[Zamora-Reina et~al.(2022)Zamora-Reina, Bravo-Marquez, and
  Schlechtweg]{Zamora2022lscd}
F.~D. Zamora-Reina, F.~Bravo-Marquez, and D.~Schlechtweg.
\newblock {LSCDiscovery}: A shared task on semantic change discovery and
  detection in {S}panish.
\newblock In \emph{Proceedings of the 3rd International Workshop on
  Computational Approaches to Historical Language Change}, Dublin, Ireland,
  2022.
\newblock URL \url{https://aclanthology.org/2022.lchange-1.16/}.

\end{thebibliography}

\end{document}